\documentclass{article}
\usepackage{graphicx} % Required for inserting images
\usepackage{amsmath, fullpage}
\usepackage{amsfonts}
\usepackage{xcolor}
\usepackage{cleveref}
\usepackage{algpseudocode}
\usepackage{float}
\usepackage{algorithm} 
\usepackage{algpseudocode}
\usepackage{booktabs}
\usepackage{natbib}
\usepackage{enumitem}
\usepackage{wrapfig}
\usepackage{url}

\title{A Technical Exploration of Causal Inference with Hybrid LLM Synthetic Data}

\author{Dana Kim\thanks{\texttt{dnalyk99@berkeley.edu}}, Yichen Xu\thanks{\texttt{yichen\_xu@berkeley.edu}}, Tiffany Lin}
\date{}

\begin{document}

\maketitle

\begin{abstract}
Large Language Models (LLMs) offer a flexible means to generate synthetic tabular data, yet existing approaches often fail to preserve key causal parameters such as the average treatment effect (ATE). In this technical exploration, we first demonstrate that state-of-the-art synthetic data generators—both GAN- and LLM-based—can achieve high predictive fidelity while substantially misestimating causal effects. To address this gap, we propose a hybrid generation framework that combines model-based covariate synthesis (monitored via distance-to-closest-record filtering) with separately learned propensity and outcome models, thereby ensuring that $(W,A,Y)$ triplets retain their underlying causal structure. We further introduce a synthetic pairing strategy to mitigate positivity violations and a realistic evaluation protocol that leverages unlimited synthetic samples to benchmark traditional estimators (IPTW, AIPW, substitution) under complex covariate distributions. This work lays the groundwork for LLM-powered data pipelines that support robust causal analysis. Our code is available at \url{https://github.com/Xyc-arch/llm-synthetic-for-causal-inference.git}.
\end{abstract}

\section{Introduction}

The growing capabilities of Large Language Models present new opportunities to support complex analytical tasks. This work introduces an LLM agent designed to enhance causal inference, not by estimating effects directly, but by addressing key data challenges that underpin causal analysis. While the causal inference community is well aware of the limitations of real-world data, the use of synthetic data to address these challenges remains under-explored. We propose that the LLM agent’s primary role is to generate high-quality synthetic data tailored for use with standard causal estimators. This synthetic data addresses issues such as privacy constraints, data scarcity, poor covariate balance, and limited overlap, while also enabling consistent benchmarking to guide estimator selection. A key challenge is ensuring that the synthetic data retains both the observational patterns and the causal structure necessary for valid inference. By tackling these foundational problems, the agent supports more reliable and robust causal analysis.

Our main contributions, therefore, directly support the data generation and utility functions of an LLM agent for causal inference:

\begin{itemize}
    \item We observe that existing generative models for privacy-preserving data sharing may fail to preserve causal parameters, even when they maintain strong predictive performance.
    \item We propose a hybrid data generation approach designed to preserve causal structure, particularly the average treatment effect (ATE).
    \item We introduce an algorithm that leverages synthetic data to mitigate violations of the positivity assumption in causal inference.
    \item We propose a method for evaluating the performance of causal estimators in scenarios where the ground truth is unknown in the real dataset.
\end{itemize}

\section{Related works}

Few studies have investigated the intersection of synthetic data generation and causal inference. Our work builds upon advances in both domains.

CTGAN \cite{xu19} pioneered conditional GANs for tabular data, introducing mode-specific normalization and training-by-sampling to address data imbalance. While preserving marginal distributions, GANs often struggle with complex conditional relationships essential for causal inference. More recently, the GReaT framework \cite{borisov23} demonstrated LLMs' capability in generating high-fidelity tabular data by encoding tables as text, though its efficacy in preserving causal structures remained unexplored until now.

Several studies have begun exploring synthetic data for causal inference. \cite{debartolomeis24} used foundation models for randomized experiments, focusing on experimental efficiency rather than addressing positivity violations. \cite{liu25} surveyed LLM applications in causal inference but did not address preserving causal parameters in synthetic data—a gap our work fills.

For positivity violations, traditional approaches like trimming extreme propensity scores can lead to biased estimates. \cite{nakada25} proposed synthetic oversampling using LLMs to address data imbalance, but primarily in predictive contexts. Our synthetic counterfactual pairing directly targets positivity challenges in causal estimation.

Evaluating causal estimators typically relies on simplified simulations that fail to capture real-world complexity. Our hybrid synthetic data framework preserves both distributional complexity and causal relationships, providing a realistic testbed with known ground truth parameters. We demonstrate that synthetic data may maintain strong predictive performance while failing to preserve critical causal relationships—highlighting the need for specialized approaches to synthetic data generation in causal inference.

\section{Using LLMs as Realistic Synthetic Data Generators}

In this section, we describe how we employed a Large Language Model (LLM), specifically GPT-2, to generate realistic synthetic tabular data using the GReaT framework \cite{borisov23}. This procedure underpins our hybrid data-generation strategy, enabling the accurate simulation of covariate-treatment-outcome triplets suitable for causal inference analysis.

\begin{wrapfigure}{r}{0.5\textwidth}
  \vspace{-10pt}              % tuck it up a bit
  \centering
  \includegraphics[width=0.48\textwidth]{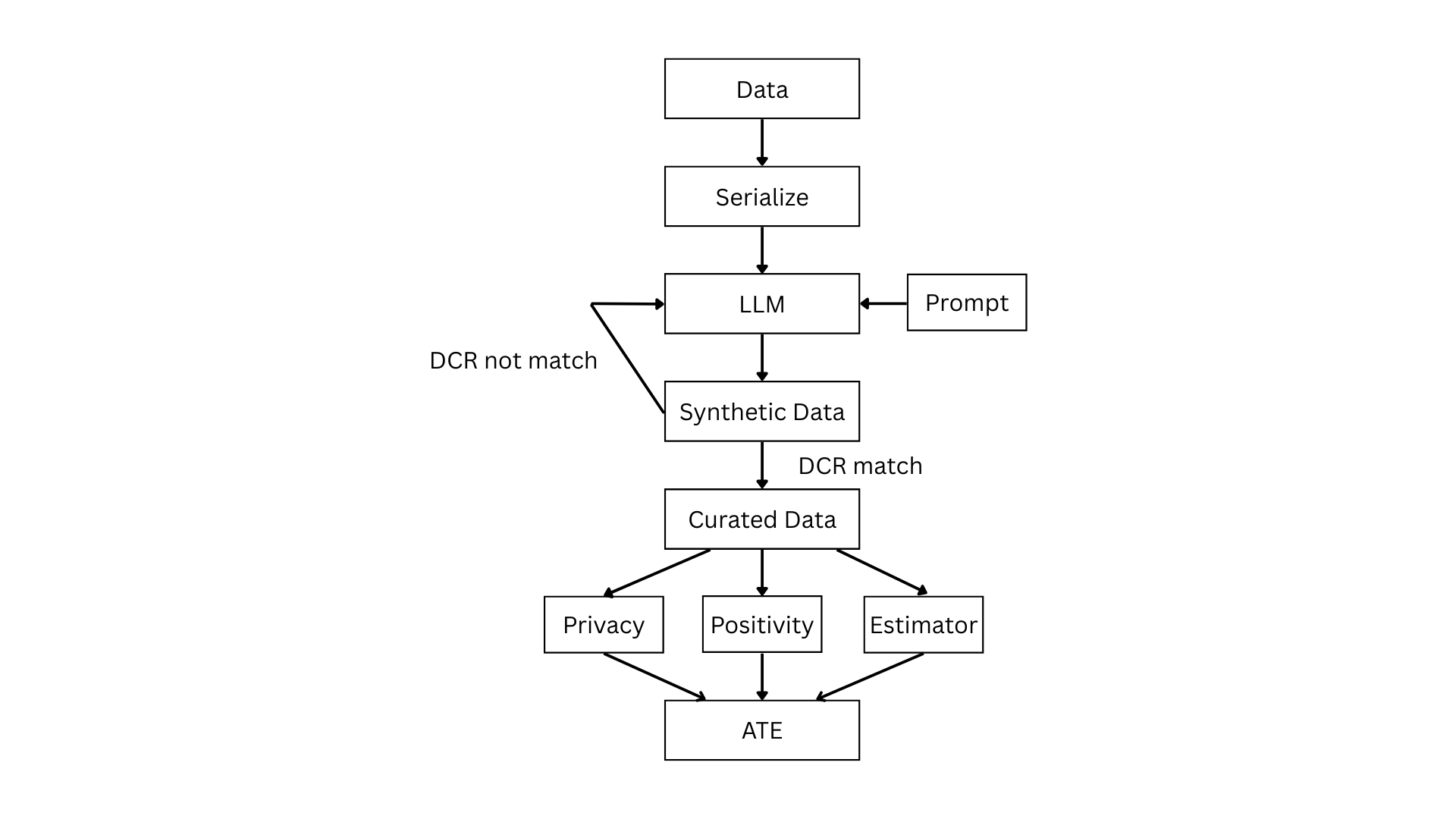}
  \caption{Overview of the LLM‐based synthetic data generation pipeline. Raw data is first serialized into prompts and processed through an LLM. The resulting synthetic samples are filtered via DCR (distance to closest record) matching. Curated synthetic data is evaluated for privacy preservation, positivity balance, and estimator fidelity before contributing to ATE estimation.}
  \label{fig:llm_pipeline}
  \vspace{-10pt}              % tighten below
\end{wrapfigure}

\subsection{Textual Serialization and Prompt Construction}

We serialized each row of the tabular dataset into natural language sentences following the methodology proposed in the GReaT framework. Each covariate, treatment, and outcome was encoded explicitly as a subject-predicate-object clause, structured as follows:
\begin{quote}
\texttt{W1=1, W2=0, W3=1, W4=0, W5=1, W6=0 => A=1, Y=0}
\end{quote}
This textual encoding preserves the semantics of each feature, enabling the LLM to model complex dependencies among covariates, treatments, and outcomes without imposing artificial ordering biases. To enhance generalizability and allow conditioning on arbitrary feature subsets, we randomized the order of serialized features prior to training.

\subsection{Fine-Tuning with GReaT}

We fine-tuned GPT-2 using the GReaT Python package \cite{borisov23}, which facilitates the application of pretrained language models for generating synthetic tabular data via textual encodings. We conducted training for 50 epochs with a batch size of 32, configuring the model to save checkpoints every 400,000 training steps using the \texttt{save\_steps} parameter to ensure reproducibility and enable checkpoint recovery. Our training dataset comprised 5,740 samples, each containing six binary covariates ($W$), a binary treatment ($A$), and a binary outcome ($Y$). The fine-tuning configuration was as follows: \textbf{Model:} GPT-2 (355M parameters); \textbf{Epochs:} 50; \textbf{Batch size:} 32; \textbf{Synthetic samples:} 50 000; \textbf{Max sampling length:} 1024–2000 tokens. After completing the training phase, we saved and subsequently reloaded the trained model to generate synthetic data. We exported the generated data in a structured tabular format to facilitate downstream causal analysis. All intermediate and final outputs follow a systematic file-naming convention to enhance traceability and reproducibility and are accessible in the supplementary repository.

\subsection{Sampling Strategy}

We leveraged the autoregressive capabilities of GPT-2 to generate synthetic records sequentially, conditioned on randomized feature orders. We conducted sampling with a maximum generation length ranging from 1024 to 2000 tokens. The flexibility inherent in the GReaT framework permitted both complete record generation and conditional sampling based on subsets of features, thereby facilitating advanced applications such as counterfactual simulations and covariate imputations.

\subsection{Advantages and Fidelity Considerations}

The GReaT framework supports arbitrary conditioning and obviates the need for lossy preprocessing methods, such as one-hot encoding. This feature is particularly beneficial for causal inference tasks, where maintaining complex conditional dependencies is critical. Furthermore, distance-to-closest-record (DCR) analysis confirmed that the generated samples closely matched the original data distribution, while simultaneously preserving sufficient variability to prevent overfitting and data leakage.

\subsection{Runtime and Reproducibility}

Across multiple runs, fine-tuning and sampling procedures averaged between 15 and 20 minutes, depending on system specifications. Our pipeline ensures reproducibility through comprehensive storage of model checkpoints and deterministic sampling configurations.

\section{Failure of Generative Model to Preserve Causal Parameter and A Hybrid generation approach}

Generative model-based data encryption using synthetic data generation to protect privacy has gained increasing attention, with approaches based on GANs \cite{xu19} and LLMs \cite{borisov23}. While metrics such as train-on-synthetic-test-on-real (TSTR) are commonly used to evaluate the predictive quality of synthetic data, there is limited focus on whether these data preserve causal parameters. The most commonly studied causal parameter is the average treatment effect (ATE), denoted as $\mathbb{E}[Y^1 - Y^0]$.

To preserve the causal parameters of a dataset beyond mere predictive performance, we focus on the tuple $(W, A, Y)$, where $W$ denotes covariates, $A$ is a binary treatment, and $Y$ is a binary outcome. Our goal is to assess whether generative models can preserve underlying causal relationships, not just surface-level predictive associations. Under standard causal assumptions (e.g., ignorability, positivity, and consistency), the average treatment effect (ATE) is identified as

\begin{align*}
    \Psi(P) = \mathbb{E}_P \big[ \mathbb{E}_P[Y \mid A = 1, W] - \mathbb{E}_P[Y \mid A = 0, W] \big].
\end{align*}

Crucially, this identification relies on how $Y$ varies with $A$ conditional on $W$. Therefore, accurate modeling of the conditional distribution $Y \mid A, W$ does not necessarily imply accurate recovery of the causal parameter.

\begin{figure}[t!]
    \centering
    \includegraphics[width=0.8\textwidth]{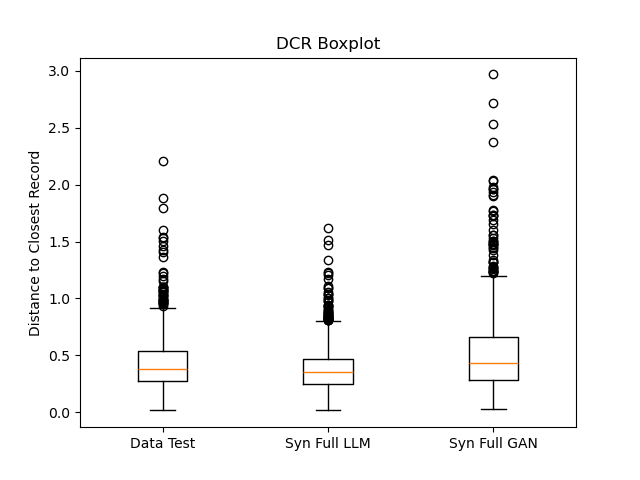}
    \caption{Boxplot of Distance to Closest Record (DCR).}
    \label{fig:dcr}
\end{figure}

Inspired by this, we propose a hybrid data generation approach to address the challenge that generative AI may fail to preserve causal parameters. As shown in \Cref{alg:hybrid}, we begin by training a generative model on the seed dataset $\mathcal{D}_{\text{seed}}$ to produce synthetic covariates $\tilde{W}$, using the distance to closest records (DCR) as a monitoring metric. Next, we fit a propensity score model $\hat{g}(A \mid W)$ and an outcome model $\hat{h}(Y \mid A, W)$ using a predictive algorithm such as random forest, both trained on $\mathcal{D}_{\text{seed}}$. Finally, for each synthetic covariate $\tilde{W}$, we sample a treatment $\tilde{A}$ from $\hat{g}$ and generate an outcome $\tilde{Y}$ from $\hat{h}$, thereby forming the synthetic triplet $(\tilde{W}, \tilde{A}, \tilde{Y})$. \Cref{fig:dcr} shows that both LLM- and GAN-based synthetic covariate generation can be well controlled by matching the DCR with that of real data (Data Test is split from real data).

\begin{algorithm}[H]
\caption{Hybrid Synthetic Data Generation Preserving Causal Structure}
\label{alg:hybrid}
\begin{algorithmic}[1]
\State \textbf{Input:} Seed dataset $\mathcal{D}_{\text{seed}} = \{(W_i, A_i, Y_i)\}_{i=1}^n$
\State \textbf{Train:}
    \Statex \quad - Train generative model $p_\theta(W)$ on $\{W_i\}_{i=1}^n$, using DCR as a monitoring metric
    \Statex \quad - Propensity model $\hat{g}(A \mid W)$ on $\mathcal{D}_{\text{seed}}$
    \Statex \quad - Outcome model $\hat{h}(Y \mid A, W)$ on $\mathcal{D}_{\text{seed}}$
\For{$i = 1$ to $n$}
    \State Sample $\tilde{W}_i \sim p_\theta(W)$ \Comment{Generate synthetic covariate}
    \State Compute $\hat{g}_i = \hat{g}(1 \mid \tilde{W}_i)$ \Comment{Propensity score}
    \State Sample $u \sim \text{Uniform}(0, 1)$
    \If{$u < \hat{g}_i$}
        \State Set $\tilde{A}_i = 1$
    \Else
        \State Set $\tilde{A}_i = 0$
    \EndIf
    \State Generate $\tilde{Y}_i = \hat{h}(\tilde{A}_i, \tilde{W}_i)$
\EndFor
\State \textbf{Return:} Synthetic dataset $\{(\tilde{W}_i, \tilde{A}_i, \tilde{Y}_i)\}_{i=1}^n$
\end{algorithmic}
\end{algorithm}

We benchmark synthetic data generated by large language models (LLMs), generative adversarial networks (GANs), and their corresponding hybrid variants. Our findings reveal that current generative AI methods for tabular data synthesis can fail to preserve key causal parameters, such as the average treatment effect (ATE). In the fully synthetic setting, denoted as "Syn Full," both treatment $\tilde{A}$ and outcome $\tilde{Y}$ are generated directly from the generative model $p_{\theta}$. To ensure privacy while maintaining similarity to the original data distribution, the synthetic covariates $\tilde{W}$ are generated with a distance to closest records (DCR) comparable to that of the real data, as shown in \Cref{fig:dcr}. The true (theoretical) ATE in our setting is $\mathbf{0.418256}$.

As shown in \Cref{tab:auc_tstr_diff}, LLM-generated synthetic data achieves the highest TSTR performance among all synthetic data methods. However, the ATE estimates obtained using either IPTW or AIPW are close to $0.5$, which is substantially higher than the true ATE. In contrast, the hybrid approach consistently yields significantly more accurate ATE estimates compared to the fully synthetic (Syn Full) method, for both GAN- and LLM-based models.

\begin{table}[htbp]
\centering
\caption{AUC TSTR, IPTW and AIPW ATE Estimates, and Their Absolute Differences from Theoretical ATE}
\begin{tabular}{l c c c c c}
\toprule
Baseline         & AUC    & IPTW ATE & AIPW ATE & IPTW Diff & AIPW Diff \\
\midrule
Data Seed        & 0.7705 & 0.3959   & 0.3770   & 0.0223   & 0.0412   \\
GAN Syn Hybrid   & 0.7425 & 0.3538   & 0.3525   & 0.0645   & 0.0657   \\
GAN Syn Full     & 0.4583 & -0.0378  & -0.0318  & 0.4561   & 0.4501   \\
LLM Syn Hybrid   & 0.7512 & 0.4295   & 0.4205   & 0.0112   & 0.0022   \\
LLM Syn Full     & 0.7545 & 0.5067   & 0.4989   & 0.0884   & 0.0806   \\
\bottomrule
\end{tabular}
\label{tab:auc_tstr_diff}
\end{table}

\begin{figure}[t!]
    \centering
    \includegraphics[width=0.8\textwidth]{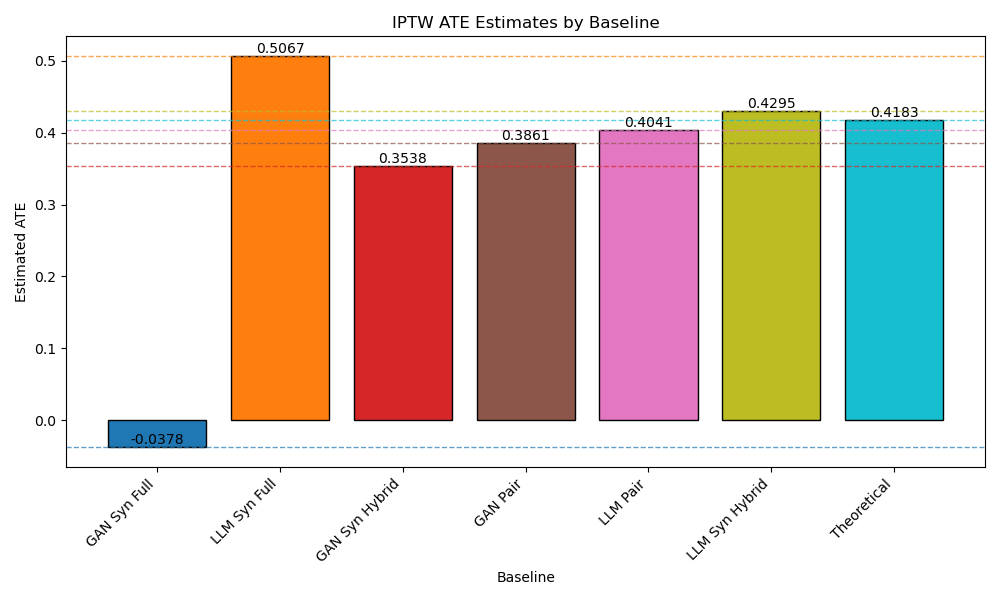}
    \caption{Bar plots of ATE IPTW estimate by different baselines.}
    \label{fig:iptw}
\end{figure}

\begin{figure}[t!]
    \centering
    \includegraphics[width=0.8\textwidth]{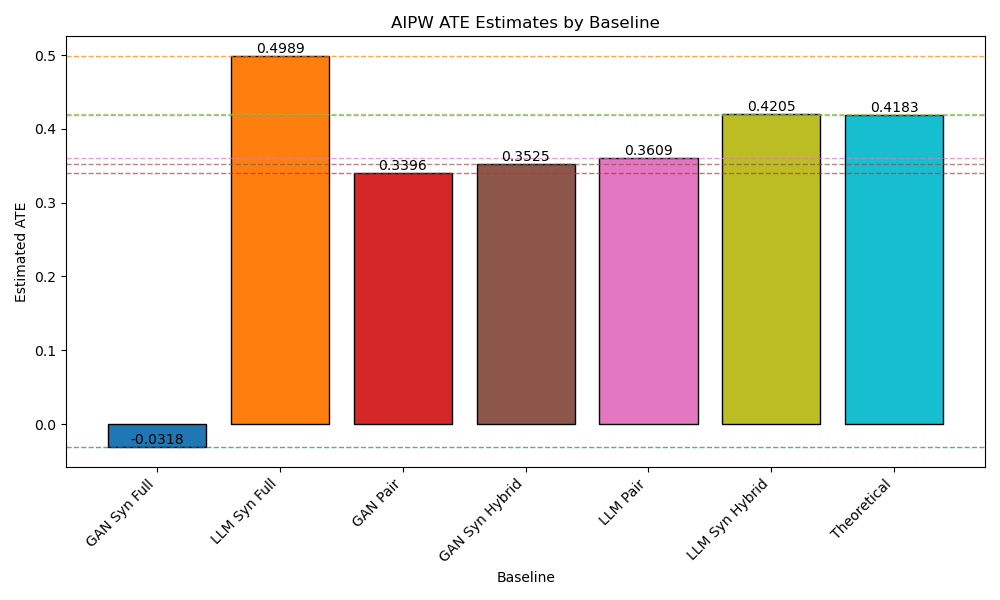}
    \caption{Bar plots of ATE AIPW estimate by different baselines.}
    \label{fig:aipw}
\end{figure}

\section{A valid way to address positivity with LLM synthetic data}

We then propose using synthetic data to address violations of the positivity assumption. Specifically, we observe that pairing samples with extreme propensity scores (PS) with synthetic counterparts (match samples by close distance bewteen covariates) can significantly reduce bias in both IPTW and AIPW estimators. In contrast, standard truncation techniques offer little benefit in this setting, as samples with near-zero propensity scores consistently receive $A = 0$. This highlights that, in such scenarios, positivity violations may be more effectively addressed through sample augmentation rather than reweighting adjustments. We define samples with propensity scores \( p_i < \frac{1}{\sqrt{n} \log n} \) as extreme PS samples. We use Euclidean distance to measure similarity. While sample pairing is a practical strategy, its debiasing effect assumes that the observed data is imbalanced, whereas the true population is not. To address this, we use synthetic data to represent the underrepresented segments of the population.

\begin{table}[htbp]
\centering
\caption{IPTW and AIPW ATE Estimates and Their Absolute Differences from Theoretical ATE}
\begin{tabular}{lccccc}
\toprule
Baseline          & IPTW ATE & AIPW ATE & IPTW Diff & AIPW Diff \\
\midrule
Raw               & 0.3169   & 0.1666   & 0.1014    & 0.2516    \\
Raw (truncated)   & 0.3169   & 0.1666   & 0.1014    & 0.2516    \\
GAN Pair          & 0.3861   & 0.3396   & 0.0321    & 0.0787    \\
LLM Pair          & 0.4041   & 0.3622   & 0.0141    & 0.0561    \\
\bottomrule
\end{tabular}
\label{tab:pos}
\end{table}

\section{LLM Synthetic data for realistic estimator evaluation}

Practitioners often struggle to validate causal estimators on real-world data due to the absence of observable ground truth. While traditional simulation-based evaluations are widely used, they are often constrained by hand-crafted, oversimplified data-generating processes and fail to capture the complexity of real-world covariate structures — particularly in high-dimensional settings.

We propose a novel evaluation framework that uses hybrid synthetic data—combining LLM-generated covariates with separately estimated treatment and outcome models—to benchmark causal estimators under realistic, scalable conditions. These datasets retain both distributional and causal structure from real data, enabling reliable estimation of bias, variance, and MSE for methods like IPTW, AIPW, and substitution.

Because synthetic data can be generated at scale, we support controlled benchmarking across many replicates while preserving real-world complexity. As shown in \Cref{tab:estimators}, our approach more closely reflects estimator performance on large real datasets than GAN-based benchmarks. Generally, both LLM- and GAN-based hybrid synthetic data approaches yield accurate proxies for $\widehat{\text{Bias}}$, $\widehat{\text{Var}}$, and $\widehat{\text{MSE}}$ relative to real data. One can generate a large synthetic dataset, estimate the effect using asymptotically unbiased methods like IPTW or AIPW (under correct model specification via, e.g., random forests), and treat this as ground truth. With unlimited synthetic samples, any estimator can then be benchmarked in this realistic, simulated environment. In our setting, the large sample size is 50,000, while each replication uses 1,000 samples.

\begin{table}[t!]
    \centering
    \caption{Large Sample Estimates (Large), approximated Bias, Variance, and MSE for IPTW, AIPW, and Simple Substitution Estimators}
    \label{tab:estimators}
    \begin{tabular}{lcccc}
        \toprule
        \multicolumn{5}{c}{\textbf{IPTW Results}} \\
        \midrule
        Baseline & Large & $\widehat{\text{Bias}}$ & $\widehat{\text{Var}}$ & $\widehat{\text{MSE}}$ \\
        \midrule
        LLM   & 0.4295 &  0.0007  & 0.000648 & 0.000649 \\
        GAN   & 0.3538 &  0.0014  & 0.000795 & 0.000797 \\
        Real & 0.4172 & -0.0019  & 0.000582 & 0.000585 \\
        \midrule
        \multicolumn{5}{c}{\textbf{AIPW Results}} \\
        \midrule
        Baseline & Large & $\widehat{\text{Bias}}$ & $\widehat{\text{Var}}$ & $\widehat{\text{MSE}}$ \\
        \midrule
        LLM   & 0.4204 & -0.0170  & 0.000635 & 0.000923 \\
        GAN   & 0.3526 & -0.0156  & 0.000739 & 0.000983 \\
        Real & 0.4137 & -0.0221  & 0.000564 & 0.001053 \\
        \midrule
        \multicolumn{5}{c}{\textbf{Simple Substitution Results}} \\
        \midrule
        Baseline & Large & $\widehat{\text{Bias}}$ & $\widehat{\text{Var}}$ & $\widehat{\text{MSE}}$ \\
        \midrule
        LLM   & 0.4190 & -0.0228  & 0.000627 & 0.001147 \\
        GAN   & 0.3511 & -0.0207  & 0.000730 & 0.001157 \\
        Real & 0.4123 & -0.0283  & 0.000559 & 0.001359 \\
        \bottomrule
    \end{tabular}
\end{table}

\section{Real-World Applications and Limitations}

We applied our method to a real-world dataset to illustrate a key limitation: without careful tuning, generative models can fail to capture essential structure, leading to synthetic data that misguides causal inference. Using NHANES data from two survey cycles (2011–2014), we built an observational dataset of ~2,200 participants, including ten covariates, a binary treatment (“cvd”), and a binary outcome (“pfq051”).

As a basic validation, we trained classifiers to predict the outcome using (a) treatment, (b) covariates, and (c) both. Covariates were highly predictive, treatment alone had moderate signal, and the combination performed best. Some covariates—like age—strongly correlated with treatment. Importantly, propensity scores showed good overlap, supporting valid causal comparisons. Nonetheless, poor tuning of generative models undermined the synthetic data’s reliability for causal analysis.

\begin{table}[t!]
\centering
\caption{Estimated ATE by IPTW, AIPW, and AUC on Test Set}
\label{tab:combined_estimates}
\begin{tabular}{lccc}
\toprule
Baseline           & IPTW ATE & AIPW ATE & AUC    \\
\midrule
Data Seed          & 0.0120   & 0.0085   & 0.8505 \\
LLM Syn Full       & 0.0075   & 0.0053   & 0.4496 \\
LLM Syn Hybrid     & –0.0004  & 0.0041   & 0.8024 \\
GAN Syn Full       & 0.0040   & 0.0035   & 0.5117 \\
GAN Syn Hybrid     & 0.0056   & 0.0118   & 0.8045 \\
\bottomrule
\end{tabular}
\end{table}

As shown in \Cref{tab:combined_estimates}, neither the LLM- nor GAN-based methods achieve a satisfactory AUC, suggesting possible issues with hyperparameter tuning. However, due to limited computational resources, we did not increase the number of training epochs. Despite the suboptimal distributional fidelity, the Hybrid method substantially improves AUC, potentially enhancing the estimation quality on GAN-generated synthetic data and aligning it more closely with the Data Seed benchmark. Notably, the estimated effect remains consistently small, providing no evidence of a significant treatment effect. Positivity violations are not a concern in this setting; hence, pairing does not offer additional benefits. This real-world scenario underscores the importance of careful tuning in generative models to ensure even coarse alignment with the target data distribution.

\section{Conclusion}
We have shown that hybrid LLM‐based synthetic data generation—combining distance‐to‐closest‐record filtering with separate propensity and outcome models—substantially improves preservation of causal parameters, especially the average treatment effect, over fully generative approaches.  

Key next steps include developing real‐time monitoring metrics to detect when synthetic data deviates from causal assumptions; deriving theoretical guarantees on estimator bias and variance under hybrid generation protocols; and extending our framework to support text‐modality data synthesis for richer causal analyses. These advances will be critical for certifying the reliability of LLM‐powered causal pipelines in high‐stakes settings.

\section{GitHub, Video, Slides, Info}

\begin{itemize}
    \item GitHub: \url{https://github.com/Xyc-arch/llm-synthetic-for-causal-inference.git}
    \item Slides: \url{https://drive.google.com/file/d/1cERbLJOYuJiHI9zeqLtNEVaawVATMHI4/view?usp=drive_link}
    \item Video: \url{https://drive.google.com/file/d/1-V3x3eGpCjH_S1MKRGfc6MsSkueeldEg/view?usp=drive_link}
\end{itemize}

\bibliography{cite}
\bibliographystyle{apalike}

\end{document}